# Mitigating Knowledge Conflicts in Language Model-Driven Question Answering

Han Cao*
University of California, San Diego
San Diego, CA, US
*Corresponding author
caoh16@gmail.com

Zhaoyang Zhang
University of California, San Diego
San Diego, CA, US
zhaoyangz2020@gmail.com

Xiangtian Li
University of California, San Diego
San Diego, CA, US
xil160@ucsd.edu

Chufan Wu
University of California, San Diego
San Diego, CA, US
chufanwu15@gmail.com

Hansong Zhang
University of California, San Diego
San Diego, CA, US
haz064@ucsd.edu

Wenqing Zhang
Washington University in St. Louis
St. Louis, MO, US
wenqing.zhang@wustl.edu

*Abstract*—In the context of knowledge-driven seq-to-seq generation tasks, such as document-based question answering and document summarization systems, two fundamental knowledge sources play crucial roles: the inherent knowledge embedded within model parameters and the external knowledge obtained through context. Recent studies revealed a significant challenge: when there exists a misalignment between the model's inherent knowledge and the ground truth answers in training data, the system may exhibit problematic behaviors during inference, such as ignoring input context, or generating unfaithful content. Our investigation proposes a strategy to minimize hallucination by building explicit connection between source inputs and generated outputs. We specifically target a common hallucination pattern in question answering, examining how the correspondence between entities and their contexts during model training influences the system's performance at inference time.

## I  Introduction

Large pre-trained language models are a common building blocks for a wide array of real world language tasks. Language models are known to encode factual knowledge during training. Such property have been exploited in tasks such as closed-book question answering and commonsense reasoning [1]. However, recent study shows that improperly activated parametric knowledge during inference could cause the model to produce factually inconsistent outputs. Such inconsistency is undesirable in natural language generation (NLG) tasks that requires grounding with respect to certain contextual information, affecting performance in various tasks such as table-to-text generation [2], abstract summarization, and grounded dialogue systems.

Large language models (LLMs) have been extremely popular for questions answering nowadays [3], with tons of engineering work done to improve its efficiency, like quantization [4] and cloud optimization [5]. The work also deeply influenced other generative AI areas like stylistic image generation. However, accuracy and consistency still remain a huge challenge for everyone.

A particular challenge in ensuring factual consistency in grounded NLG tasks is that the usually unsatisfied sufficiency between the grounding context and gold output. In the ideal world, the input context should always contain enough information for a model to produce the gold output. However, such an assumption is impractical in many tasks. For example, human-written gold answers in summarization often contains extrinsic hallucination, where extra background knowledge is required for the model to produce the correct output. Similarly, the one-to-many problem in dialogue system were also caused by the insufficiency to accurately predict response given contextual information. Thus, it is crucial that we develop training scheme to constrain model behavior under such sufficiency constraint.

We study a simple variant of hallucination, entity-based knowledge conflicts [6]. Concretely, recent study find reader model exhibits overreliance on parametric knowledge, primarily due to information insufficiency caused by an inperfect retriever during training. In other words, the reader model learns to ignore retrieved documents and rely on parametric knowledge during training, resulting in hallucination at test time. While such behavior can be beneficial, un-alerted hallucination significantly affects model usability and trustworthiness. We propose two sets of methods to mitigate undesirable model memorization. We hope to show that with minimum performance sacrifice, our model generate output that is more faithful with respect to input context.

## II  Background and Related Works

### A.  Hallucination in Text Generation

As mentioned in the introduction though, some early studies focused on the potential pitfalls of leveraging standard likelihood maximization-based objectives in train-ing and decoding Natural Language Generation models. They found that an approach that maximizes this potential could lead to deterioration. At the same time, it turns out that these models often produce meaningless text or are not faithful to the source input. Researchers have come to call this unwelcome generation a hallucination. Hallucination in language models poses significant risks, including potential privacy violations. Carlini et al. [7] demonstrated that language models can be deliberately prompted to recall and generate sensitive personal information embedded within the training data. This phenomenon, characterized by the memorization and retrieval of training data, is classified as a form of hallucination [8].

Many LLMs are evolving from text-only to multimodal content understanding and question answering, which proposes a bigger challenge for multimodal understanding [9]. Previous purely dialog-based models like[10] intend to capture more structural information instead of knowledge. Researchers are also utilizing Transformer and other related models for visual tasks [11] and medical tasks [12]. These applications introduce higher requirements for knowledge conflict resolution.

Given that question-answering (QA) models rely on external knowledge to retrieve information relevant to a given question and generate answers accordingly, a critical objective is to ensure that answers are grounded in the retrieved documents. Hallucinations in the generated answers can significantly mislead users and severely compromise the model's overall performance. [13].

### B. Controllable Generation

Controllable text generation [14] may be one tool to resolve the hallucination problem. Unlike plain text generation, we want the sentences generated by the models ($x$) to align with specific attributes ($a$), such as sentiment and content. Works in this area typically deal around the conditional probability $p(x|a)$. While some authors [15] decide to directly model it, some avoid direct modeling through bayes rules or latent variables. For example, Hu et al. uses a latent variable ($z$) to transform the problem into modeling $p(x|z,a)$, $p(a|x)$ and $p(z|x)$ separately. Such Bayesbased methods have been widely adapted in areas like casual inference [16] and operation research [17]. This tool may allow us to treat the easily hallucinated entities as attributes and consequently train the model not to completely ignore the provided context [18].

### C. Question Answering

Question answering (QA) is a major research direction of NLP. It has broad applications in search engine, intelligent assistants, and education [19]. Two of the largest subbranches under QA are visual question answering [20][21], which uses images as sources of knowledge and context, and text question answering [22], which uses text as sources, or multimodal [23]. One of major subbranches under text question answering is called open-domain question answering [24]. In this task, given any question, the model is expected to find the answer to it from a giant database of passages, e.g. Wikipedia articles. The state-of-the-art is the retriever-reader approach, as mentioned in Chen et al [25]. With this approach, first, a retriever, such as a dense retriever or simply a BM25, will first retrieve a small set of relevant passage from the given database of passages [26]. A document reader, typically a standard language model like RNN [27] or BERT, will then read through these retrieved passages and produce the answer to the question.

## III. METHODS

This section is organized as follows: we first cover definition of knowledge conflicts and a simple data augmentation baseline solution. Then we introduce our projected directions that could potentially lead to better solution to knowledge conflicts. Finally, we report our tentative schedule for the project.

### A. Problem re-formalization and Model selection

Seq2seq-based big QA models like BART, when accompanied by QA datasets with contexts, become hard to train and require computing resource beyond our current constraints [28]. Hence, to seamlessly handle the datasets we used in our experiments, which typically contain long contexts with more than 150 words, we used pretrained generative language models and finetuned them for QA tasks.

Similar to what is done in most language models, question and answer pairs are connected by a special token for training, and model are only given questions as prefixs for inference. Consequently, this task can be viewed as a conditional text generation problem. Given this, the goal is to optimize the probability that a language model $\mathcal{M}$ generates the correct entity answer $x$ given a query $q$ and a context $c$, we formalize the task as maximizing the conditional probability:

$$P(x \mid q, c, \mathcal{M})$$

To achieve this, we optimize the model parameters θ such that:

$$\theta^* = argmax_\theta E_{(q,c,x) \sim D}[log P_\theta(x \mid q, c)]$$

- $D$ is the training dataset containing examples of queries, contexts, and correct answers.
- $P_\theta(x \mid q, c)$ is the probability of generating x conditioned on q and c, parameterized by θ.
- We optimize $\theta$ in autoregressive way like language modeling

This process can be described as stage 1 in Figure 1.

### B. Knowledge Conflicts and Baseline Method

In open-book question answering, the reader model is trained to produce an answer $x$ given a (retrieved) context $c$ and a query $q$. Ideally, the reader model should learn to always estimate the answer given *both* the context and the query, i.e. $p_\theta(x|q,c)$. However, previous study found that via factual knowledge encoded in model parameters, the reader model sometimes ignores the context, and directly estimate the answer, i.e. $p_\theta(x|q)$. We believe a straightforward solution to this problem is simply by training on a set of augmented context-query-answer triplets of the form $\{x', q, c'\}$, where $x'$ is only answerable using information from $c'$. However, this method requires heuristic based design of augmented data, and requires fine tuning the entire model parameters, thus is computationally and labor expensive [29].

Thus, instead we can build anti-factual dataset $\{x', q, c'\}$, where $x'$ is substitution of $x$ with same entity type, and $c'$ is simply generated by replace all x in $c$ with x′ .

To evaluate the memorization rate of a language model, we define the memorization rate Rmem as:

$$R_{mem} = \frac{1}{|\mathcal{D}'|} \sum_{(q,c',x') \in \mathcal{D}'} \mathbb{I}[arg\ \max_{x''} P_\theta(x'' \mid q, c') = x]$$

where:

- $|\mathcal{D}'|$ is the new anti-factual dataset.
- Based on the specified condition, $\mathbb{I}[\cdot]$ equal 1 when condition is True and 0 when condition is False.

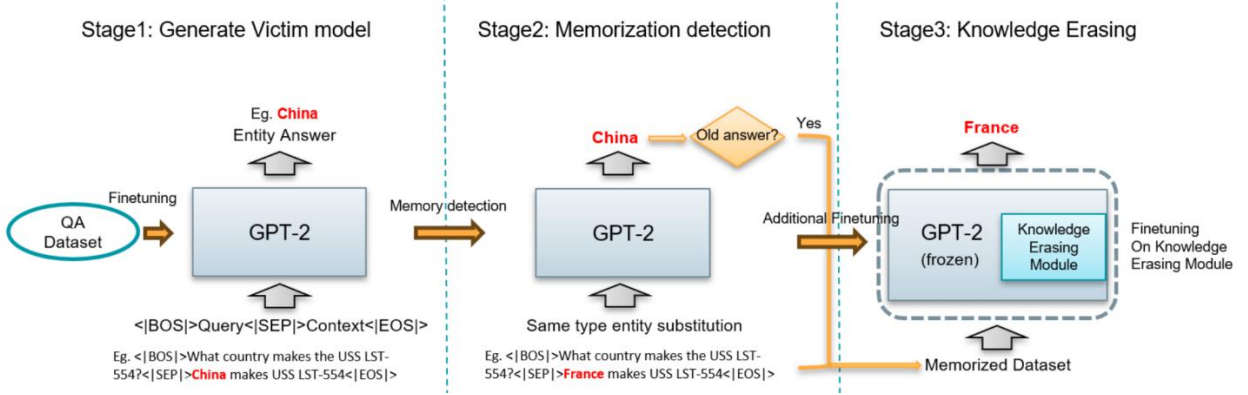

FIG 1 THREE STAGES OF MODEL FIRST FINETUNE LM ON QA TASK TO GENERATE VICTIM MODEL; SECOND CALCULATE MEMORIZATION RATE TO GENERATE MEMORIZED DATASET; THIRD, ERASE KNOWLEDGE BY ADDITIONAL MODULE FINETUNING

- $P_\theta(x'' \mid q, c')$ is the probability assigned by the model to generate answers $x''$.
- $x$ is the memorized answer associated with $q, c$ in the previous training set.

This process can be described as stage 2 in Figure 1.

*C. Knowledge Erasing Module*

We already know knowledge can be stored in language models and this attribute has been utilized in various areas including chat robots and network security. However, erasing or modification of the stored knowledge remains a challenge.

Previous work demonstrate that additional trainable parameters could be used to infuse factual knowledge into pretrained language models [30]. In this work, we investigate whether the same formulation could be used to steer the model towards forgetting memorized undesirable correlation. Namely, we investigate two types of extra parameters: bottle neck adapter [31] and prefix tuning. The former closely resembles the knowledge infusing module used by [30], and the latter is closed connected to universal trigger attack, which is known to be able to steer model behavior. Besides, inspired the adversarial loss used by [32], we trained our adapters in adversarial way for robust performance.

Now our optimization process can be represented as following:

$$\theta_{add}^* = arg \max_{\theta_{add}} \frac{1}{N} \sum_{i=1}^{N} \mathcal{L}(x_i, y_i; \theta_M, \theta_{add})$$

$$\theta_M = fixed\ weights$$

Among them

- $\mathcal{L}(x_i, y_i; \theta_M, \theta_{add})$ stands for our cross-entropy loss same as language modeling training
- $\theta_M$ is fixed language model parameters
- $\theta_{add}$ is knowledge erasing module parameters

This process can be described as stage 3 in Figure 1.

The benefit of using extra parameter are three folds: firstly, the small amount of additional parameters do not require the original model parameters to be finetuned, thus alleviates the concern of catastrophic forgetting; secondly, the incorporated extra parameter provides potential future experiment for analyzing how parameterized knowledge changes during finetuing; thirdly, small adapters are parameter-efficient and makes finetuning significantly faster [33].

According to recent work on LLM reasoning [34], we are still limited in the ability of LLM reasoning capacity. Previously reinforcement learning-based methods have been explored to align machine learning models with human behavior [35]. Recent work has been done to enhance this by space exploration [36]. Though it's still at the beginning stage, it's an important direction to explain how LLM understands knowledge.

IV  DATASETS

*A. KMIR Dataset*

a) Introduction to KMIR: For preliminary experiments, we used the recently released KMIR dataset from the study [37], which is a new benchmark for evaluating knowledge memorization, identification and reasoning for language models. It includes questions and answers for different types of knowledge, such as commonsense, general knowledge and domain specific knowledge. This dataset is more light-weighted.

b) Preprocessing: Here we only use the dataset for knowledge memorization evaluation. Thus we only selected the data with Triple Completion relation type. Then for each query statement, we generated a question accordingly via a huggingface api function. In this way we got more than 140,000 sample for training.

*B. Natural Question(NQ) Dataset*

As mentioned in [38], this study proposed a extensible framework to generated entity-based knowledge conflicts QA datasets. They focused on 5 types of entities including PER, DAT, NUM, ORG, LOC, which stand for person, date, numeric, organization, and location accordingly. They also introduced four types of substitutions to comprehensively understand model behavior: corpus substitution, alias substitution, type swap substitution and popularity substitution.

We can follow this former work and use [39]'s Natural Questions (NQ) [40]) for our experiments. We only focused on the corpus substitution setting because it replaced the answer with another answer of the same type, which aligned with the setting before. This type of substitution also made it easier to expect reasonable answers from the model.

While the framework also allows custom substitutions. Inspired by the framework of [41], for future work, we can further create more types of entity substitution policies to enhance our model's ability to reduce hallucination. Besides, more knowledge bases (KBs) like YAGO can be inte- grated into this framework to increase the diversity of potential entity candidates for substitution [42].

### C. Victim Model

We used pretrained GPT-2 for the task. The questions and answers are combined with special tokens to show the model the pattern. Then we finetuned GPT2 on the training dataset for 2-3 epochs. To quantify the memorization of our model, we then used the finetuned model to generate answers given the questions. If the model generated the exact same answer by comparing the texts, we considered the model memorized the question and answer pair.

We also considered the situation that which the model mentioned the correct entity in the answer but described it in a different way. This situation did happen. But this may also introduce some perturbed answers with hallucination. Therefore, we strictly counted the exactly matched answers only.

*a) For KMIR Datset:* We randomly sampled 4,000 samples from the training set to evaluate the memorization rate. The results showed that the exactly matched rate is 12%-13%. While the same entity rate is about 25% to 30%. We finally got 593 memorized samples.

Then we categorized the data according to the pred type and gathered the answers within each category. We generated a wrong entity for each question by randomly selecting from according category. In this way, we made sure that the entity type is the same as original answer, so that the model can be more easily stirred to generate the wrong answer different from its memory.

*b) For NQ Dataset:* Similarly, we chose 10,000 samples from the training set and the memorization rate of model is about 5-6%. We finally got 637 memorized samples.

## V. EXPERIMENT

### A. Qualitative results for Prompt tuning

Following the prompt tuning method, we clarified at III-C, the result is as follow in Table I:

TABLE I EXAMPLES OF PROMPT TUNING ON KMIR

| Question | What is Heinz Hohner's nationality? |
|---|---|
| Context | Heinz Hohner nationality is French Navy. |
| Original Answer | Heinz Hohner nationality is Germany. |
| Bottleneck adapter Answer | Heinz Hohner nationality is France. |
| Prefix tuning adapter Answer | Heinz Hohner nationality is French Navy. |

Before shuffling, the original victim model memorized that the original answer is "Heinz Hohner nationality is Germany". We then used two types of adapters for prompt tuning, bottleneck adapter and prefix tuning adapter. By introducing the shuffled new answer as the context, we would like the model to generate the new desired answer. The model after prompt tuning successfully output the shuffled answer. Through comparison between the two adapters, we found that the bottleneck adapter may be more generative because it answered "France" instead of "French Navy".

Similarly, on the NQ dataset, the model learnt to pay attention to the context and changed its answer. There are 2 examples in the table II. As we can see, the model changed its original answer to the red-marked entity in the context, which showed that the model understood the relation between contexts and questions.

### B. Quantitative Evaluation for Prompt tuning

The evaluation metrics is now different from [38]. For our model, we focus on the questions that the model correctly answers on the original unmodified samples. Through the exact match measurement, we can compare the predictions on the original example $x$ and the example $x'$ after substitution. Then for the predictions on $x'$, we calculate the fraction of the substituted answer $p_s$ as the accuracy.

$$p_s = \frac{1}{|\mathcal{D}_{\text{test}}|} \sum_{(q,c',x') \in \mathcal{D}_{\text{test}}} \mathbb{I}\left[\hat{x} = x'\right]$$

where:

- $|\mathcal{D}_{\text{test}}|$ is the size of the test dataset.
- $\hat{x}$ is the model's predicted answer.

TABLE II EXAMPLES OF PROMPT TUNING ON NQ

| | Example1 | Example2 |
|---|---|---|
| Question | who was the season 1 winner of american idol? | when does bigg boss season 2 tamil start? |
| Original Answer | Kelly Clarkson | 2018 |
| Context | <P> The inaugural season of American Idol: The Search for a Superstar aired from June 11 to September 4, 2002, culminating in **Hugh O'Brian** being crowned the winner. Co-hosted by Ryan Seacrest and Brian Dunkleman, the | <P> Kamal Haasan hosted the debut season of Bigg Boss Tamil, which premiered on Star Vijay on June 25, 2017. Building on its success, Star Vijay announced the second season, Bigg Boss Tamil 2, in October 2017, with its |

|  | season marked Dunkleman's sole appearance, as he departed from the series following its conclusion </P> | broadcast scheduled for **August** the following year </P> |
|---|---|---|
| **Bottleneck adapter Answer** | Hugh O'Brian | August |
| **Prefix tuning adapter Answer** | Hugh O'Brian | August |

TABLE III  RESULT OF PROMPT TUNING ON KMIR

|  | accuracy on training set | accuracy on test set |
|---|---|---|
| **Bottleneck adapter** | 92.1% | 92.9% |
| **Prefix tuning adapter** | 88.0% | 91.9% |

TABLE IV  RESULT OF PROMPT TUNING ON NQ

|  | accuracy on training set | accuracy on test set |
|---|---|---|
| **Bottleneck adapter** | 90.8% | 64.1% |
| **Prefix tuning adapter** | 90.8% | 64.8% |

The accuracy then measures how often the model predicts the substituted answer in the context. This metrics shows the extend to which model pay attention to the changing information in the context.

The results for KMIR and NQ are shown in Table III and Table IV. Through comparison, it is obvious that the NQ dataset is much more difficult because it has more com-plex contexts information, which requires the understanding ability of the language model. But the model still achieved pretty good performance.

## VI  CONCLUSION

Undesired storage of parameterized knowledge in pretrained models lead to factual hallucination at inference time. In this work, we investigate two classes of prompt tuning methods: Bottleneck adapter and Prefix tuning adapter. We show these two methods could effectively override memorized parameterized knowledge.